\documentclass{article} 
\usepackage{nips13submit_e,times}
\usepackage{hyperref}
\usepackage{url}

\usepackage{color}
\usepackage{amsmath}
\usepackage{bm}
\usepackage{amsthm}

\usepackage{times}
\usepackage{graphicx} 
\usepackage{subfigure} 

\usepackage{natbib}

\usepackage{algorithm}
\usepackage{algorithmic}

\usepackage{amssymb}
\usepackage{xr}
\usepackage{lscape}

\nipsfinalcopy

\newtheorem{definition}{Definition}
\newtheorem{proposition}{Proposition}

\def\mb{\mathbb}
\def\defeq{\triangleq}
\def\mc{\mathcal}

\def\diag{\mathrm{diag}}

\def\b{\mathbf{b}}

\def\d{\mathbf{d}}

\def\f{\mathbf{f}}
\def\g{\mathbf{g}}
\def\h{\mathbf{h}}

\def\p{\mathbf{p}}

\def\v{\mathbf{v}}
\def\w{\mathbf{w}}
\def\x{\mathbf{x}}
\def\y{\mathbf{y}}

\def\D{\mathbf{D}}

\def\U{\mathbf{U}}

\def\W{\mathbf{W}}
\def\X{\mathbf{X}}

\allowdisplaybreaks

\title{A Primal-Dual Method for Training Recurrent Neural Networks Constrained by the Echo-State Property}

\author{
Jianshu Chen
\\
Department of Electrical Engineering\\
University of California\\
Los Angeles, CA 90034, USA \\
\texttt{cjs09@ucla.edu} \\
\And
Li Deng \\
Machine Learning Group\\ Microsoft Research\\
Redmond, WA 98052, USA  \\
\texttt{deng@microsoft.com} \\
}

\begin{document}

\maketitle

\begin{abstract} 

We present an architecture of a recurrent neural network (RNN) with a fully-connected deep neural network (DNN) as its feature extractor. The RNN is equipped with both causal temporal prediction and non-causal look-ahead, via auto-regression (AR) and moving-average (MA), respectively. The focus of this paper is a primal-dual training method that formulates the learning of the RNN as a formal optimization problem with an inequality constraint that provides a sufficient condition for the stability of the network dynamics. Experimental results demonstrate the effectiveness of this new method, which achieves 18.86\% phone recognition error on the TIMIT benchmark for the core test set. The result approaches the best result of 17.7\%, which was obtained by using RNN with long short-term memory (LSTM). The results also show that the proposed primal-dual training method produces lower recognition errors than the popular RNN methods developed earlier based on the carefully tuned threshold parameter that heuristically prevents the gradient from exploding. 

\end{abstract} 

\section{Introduction}
\label{Sec:Intro}

Considerable impact in speech recognition has been created in recent years using  fully-connected deep neural networks (DNN) that drastically cut errors in large vocabulary speech recognition \citep{DongDengDahl,dahl2011large,seide2011conversational,Dahl2012,kingsbury2012scalable,hinton2012deep,deng2013new,DengMicrosoft,Yu2013,dahl2013improving,sainath2013optimization}. However, the well-known problem of the previous state-of-the-art approach based on Gaussian Mixture Model (GMM)-HMMs has not been addressed by the DNNs in a principled way: missing temporal correlation structure that is prevalent in the speech sequence data. Recurrent neural networks (RNN) have shown their potential to address this problem \citep{robinson1994application,graves2013speech}, but the difficulty of learning RNNs due to vanishing or exploding gradients \citep{pascanu2012difficulty} or the complexity of the LSTM (long short-term memory) structure in RNNs \citep{graves2013speech} have so far slowed down the research progress in using RNNs to improve speech recognition and other sequence processing tasks.

In this paper, we propose an architecture of RNNs for supervised learning, where the input sequence to the RNN is computed by an independent feature extractor using a fully-connected DNN receiving its input from raw data sequences. We have formulated both autoregressive (AR) and autoregressive and moving-average (ARMA) versions of this RNN.  A new learning method is developed in this work, which is successfully applied to both AR and ARMA versions of the RNN. The new method frames the RNN learning problem as a constrained optimization one, where cross entropy is maximized subject to having the infinity norm of the recurrent matrix of the RNN to be less than a fixed value that provides a sufficient condition for the stability of RNN dynamics.  A primal-dual technique is devised to solve the resulting constrained optimization problem in a principled way. Experimental results on phone recognition demonstrate: 1) the primal-dual technique is effective in learning RNNs, with satisfactory performance on the TIMIT benchmark; 2) The ARMA version of the RNN produces higher recognition accuracy than the traditional AR version; 3) The use of a DNN to compute high-level features of speech data to feed into the RNN gives much higher accuracy than without using the DNN; and 4) The accuracy drops progressively as the DNN features are extracted from higher to lower hidden layers of the DNN.

\section{Related Work}
\label{Sec:RelatedWork}

The use of recurrent or temporally predictive forms of neural networks for speech recognition dated back to early 1990's \citep{robinson1994application,deng1994analysis}, with relatively low accuracy.
Since deep learning became popular in recent years, much more research has been devoted to the RNN \citep{graves2006connectionist,graves2012sequence,maas2012recurrent,mikolov2012statistical,vinyals2012revisiting,graves2013speech}.
Most work on RNNs made use of the method of Back Propagation Through Time (BPTT) to train the RNNs, and empirical tricks need to be exploited in order to make the training effective. The most notable trick is to truncate gradients when they become too large
\citep{mikolov2011strategies,pascanu2012difficulty,bengio2013advances}. Likewise, another empirical method, based on a regularization term that represents a ``preference for parameter values,'' was introduced to handle the gradient vanishing problem \citep{pascanu2012difficulty}. 

The  method we propose for training RNN in this paper is based on optimization principles, capitalizing on the cause of difficulties in BPTT analyzed in \citep{pascanu2012difficulty}. Rather than using empirical ways to scale up or scale down the gradients, we formulate the RNN learning problem as an optimization one with inequality constraints and we call for a natural, primal-dual method to solve the constrained optimization problem in a principled way \citep{poliak1987introduction}. Our work also differs from the earlier work reported in \citep{sutskever2013training,martens2011learning}, which adopted a special way of initialization using echo state networks \citep{jaeger2001echostate} when carrying out the standard BPTT procedure, without using a formal constrained optimization framework.

This work is originally motivated by the echo state network \citep{jaeger2001echostate}, which carefully hand-designed the recurrent weights making use the 
 the echo-state property. The weights are then fixed and not learned, due to the difficulty in learning them. Instead, in this work, we devise a method of learning these recurrent weights by a formal optimization framework which naturally incorporates the constraint derived from the echo-state property. The echo-state property proposed in  \citep{jaeger2001echostate} provides a sufficient condition for avoiding the exploding gradient problem. Although such a sufficient constraint looks restrictive, it allows the RNN to be trained in a relatively easy and principled way, and the trained RNN achieves satisfactory performance on the TIMIT benchmark.

As another main contribution of this paper, we build a ``deep'' version of the RNN by using an independent DNN to extract high-level features from the raw speech data, capitalizing on the well established evidence for the power of DNNs in carrying out automatic feature extraction \citep{hinton2012deep,dahl2011large,lecun2012learning}. Having the DNN as the feature extractor and the RNN as the classifier trained separately helps reduce overfitting in the training. This strategy of building a deep version of RNNs contrasts with other strategies in the literature, where the same RNN model is stacking up on top of each other and all layers of RNNs are trained jointly. Aside from using weight noise for regularization, the overfitting problem caused by joint training of many RNNs has been circumvented mainly by exploiting elaborate structures such as LSTM \citep{graves2013speech}, making the complexity of the overall model higher and the results more difficult to reproduce than our simpler approach to building the deep version of the RNN reported in this paper.

Finally, our method exploits the DNN features in a very different way than the ``tandem'' architecture discussed in \citep{Tuske12}, where posterior outputs  at the top layer of the DNN is concatenated with the acoustic features as new inputs to an HMM system. We take the hidden layer of the DNN as the features, which are shown experimentally in our work to be much better than the top-layer posterior outputs as in the tandem method. Further, rather than using the GMM-HMM as a separate sequence classifier in \citep{Tuske12}, we use the RNN as the sequence classifier.

\section{The Recurrent Neural Network}
\label{Sec:DRNN}
  
\subsection{Basic architecture}
\label{Sec:DRNN:Arch}

\begin{figure}
	\centering
	\includegraphics[width = 0.48\textwidth]{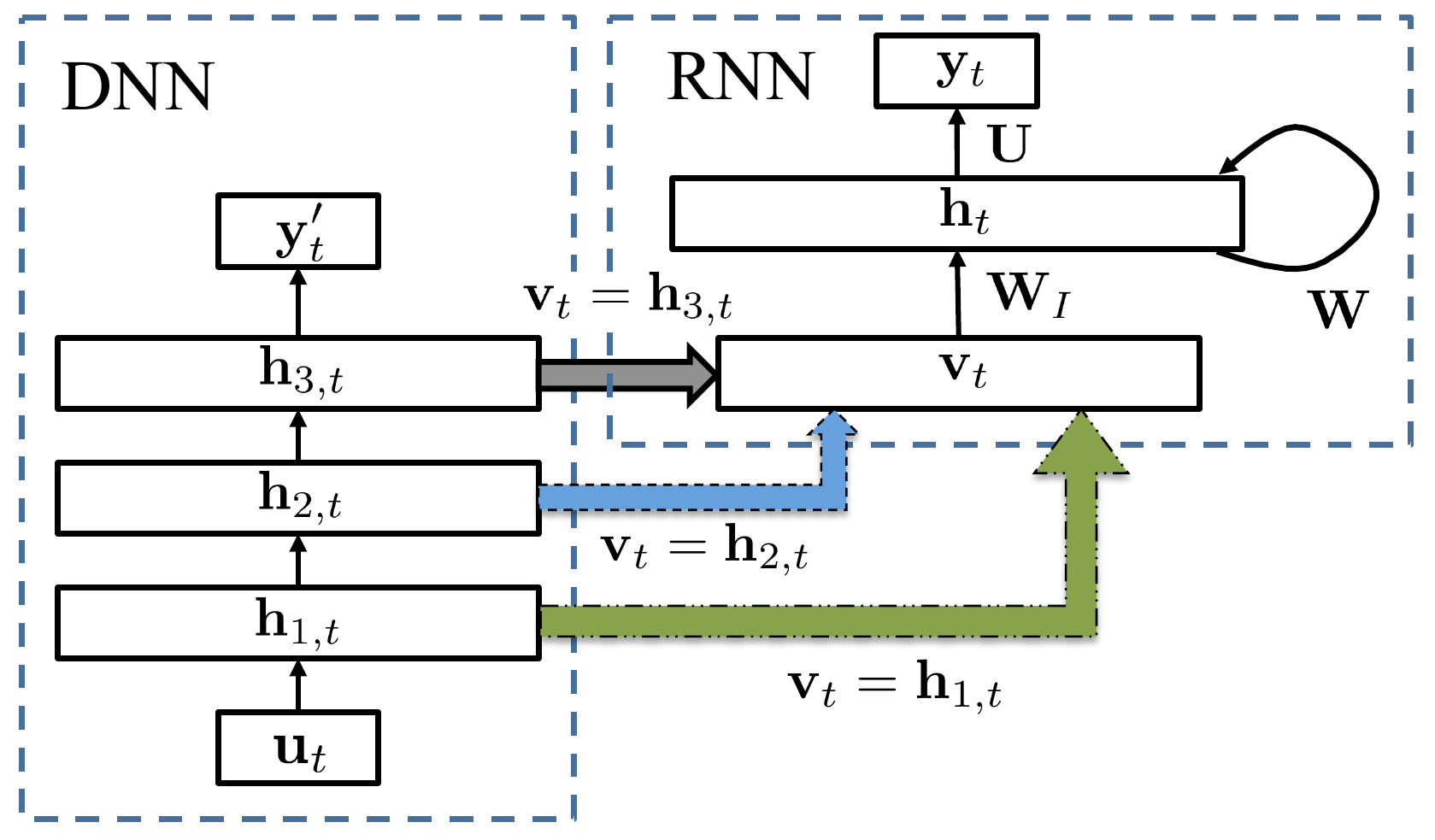}
	\caption{Architecture of the recurrent neural network with DNN features, where one of the hidden layers of the DNN will be
	used as input for the RNN. Arrows of different colors from the DNN to the RNN indicate alternative ways of 
	providing the inputs to RNN. }
	\label{Fig:DeepRNN}
\end{figure}

We consider a deep-RNN shown in Fig. \ref{Fig:DeepRNN}. $\y_t'$ denotes the output of the DNN that is trained over randomly permuted input sequence, and $\y_t$ denotes the output of RNN corresponding to the input sequence of the original order. The
network consists of two parts: (i) a lower-level DNN, and (ii) an upper-level RNN. The major task of the lower DNN
is to extract useful features from the raw data and feed them to the upper
RNN component, which replaces HMM as a sequence classifier. The RNN in the upper level captures the temporal contextual
information in its hidden states and generates the outputs that predict
the labels of the input data in the supervised sequential prediction task such as recognizing phones or words embedded in a sentence.

\subsection{Mathematical  formulation of the RNN}
\label{Sec:DRNN:Form}

In this section, we  focus on the RNN part of the deep network, given the DNN outputs that provide input vectors $\v_t$'s to the RNN. 
A standard RNN is described by the following equations:
	\begin{align}
		\h_t 	&=	\f\left(\W \h_{t-1} + \W_I \v_t + \b \right)
		\label{Equ:DRNN:Model_hiddenEqu}
					\\
		\y_t	&=	\g\left(\U \h_t \right)
		\label{Equ:DRNN:Model_outputEqu}
	\end{align}
where $\h_t \in \mb{R}^N$, $\v_t \in \mb{R}^{N_I}$ and $\y_t \in \mb{R}^{N_o}$
are vectors at discrete time $t$ that represent the hidden states, the inputs, and the outputs, respectively,
and the matrices $\W \in \mathbb{R}^{N \times N}$, $\W_I \in \mb{R}^{N \times N_I}$,
$\U \in \mb{R}^{N_o \times N}$, and vector $\b \in \mb{R}^{N}$ collect the recurrent weights, input weights, the output weights and the bias. The function $\f(\x)$ represents the nonlinearities that are applied to each entry of the vector
$\x$, i.e., the nonlinear operation that is applied to the vector component-by-component:
	\begin{align}
		\f(\x)	=	\begin{bmatrix}
						f(x_1) & \cdots	&	f(x_N)
					\end{bmatrix}^T
					\nonumber
	\end{align}
where $x_k$ denotes the $k$-entry of the vector $\x$, and each $f(x_k)$ can be sigmoid, $\tanh$ or rectified. (We will use sigmoid nonlinearity through out this work as an example.) And $\g(\x)$ represents the operation applied at the output units. Typical choices are linear function $\g(\x)=\x$ or soft-max operation.

While the standard RNN model described by \eqref{Equ:DRNN:Model_hiddenEqu} updates the hidden states $\h_t$ from its summary of the history $\h_{t-1}$ and the current input $\v_t$, which we call the autoregressive (AR) version, we can have a more general ARMA version described by
	\begin{align}
		\h_t	&=	\f\left( 
							\W \h_{t-1} 
							+ 
							\sum_{\tau = -\Delta_1}^{\Delta_2} 
							\W_{I,\tau}
							\v_{t-\tau}
							+
							\b
					\right)
		\label{Equ:DRNN:Model_hiddenEqu_contextwindow}
	\end{align}
where $\Delta_1$ and $\Delta_2$ denote the number of inputs that the network looks forward and backwards. If $\Delta_1=0$, then it only looks backwards into the past history and if $\Delta_2=0$, it only looks into future. The models described by \eqref{Equ:DRNN:Model_hiddenEqu} and \eqref{Equ:DRNN:Model_hiddenEqu_contextwindow} are parallel to the vector AR and ARMA models, respectively,  in time series analysis except that the classical AR and ARMA models are linear:
	\begin{align}
		\mathrm{[AR]:~}
		\h_t	&=	\W \h_{t-1} + \W_I \v_t					\\
		\mathrm{[ARMA]:~}
		\h_t	&=	\W \h_{t-1} + \sum_{\tau = -\Delta_1}^{\Delta_2} \W_{I,\tau} \v_{t-\tau}
	\end{align}	 
In the context of ARMA version of the RNN, $\W \h_{t-1}$ is called AR part, and the term $\sum_{\tau = -\Delta_1}^{\Delta_2} \W_{I,\tau} \v_{t-\tau}$ is called MA part. Depending on the nature of the time sequence, some tasks are easier to be modeled by AR, and some others by MA, or jointly ARMA. Just as AR model is a special case of ARMA model ($\Delta_1=\Delta_2=0$), model \eqref{Equ:DRNN:Model_hiddenEqu} is a special case of \eqref{Equ:DRNN:Model_hiddenEqu_contextwindow}. Due to its generality, the ARMA version of the RNN \eqref{Equ:DRNN:Model_hiddenEqu_contextwindow} is expected to be more powerful than the AR version \eqref{Equ:DRNN:Model_hiddenEqu}. Indeed, our experimental results presented in Section 5 have confirmed this expectation. 

A key observation we make is that in order to develop a unified learning method, \eqref{Equ:DRNN:Model_hiddenEqu_contextwindow} can be converted back to the form of \eqref{Equ:DRNN:Model_hiddenEqu} by defining some extra augmented variables. Let $\overline{\v}_t$ and $\overline{\W}_I$ be defined as
	\begin{align}
		\overline{\v}_t	&\defeq		\begin{bmatrix}
										\v_{t-\Delta_2}^T & \cdots & \v_{t+\Delta_1}^T
									\end{bmatrix}^T
								\\
		\overline{\W}_I	&\defeq		\begin{bmatrix}
										\W_{I, -\Delta_2} & \cdots & \W_{I, \Delta_1}
									\end{bmatrix}
	\end{align}
Then, \eqref{Equ:DRNN:Model_hiddenEqu_contextwindow} can be written in the following equivalent form:
	\begin{align}
		\h_t	&=	\f\left(
						\W \h_{t-1} + \overline{\W}_I \overline{\v}_t + \b
					\right)
	\end{align}
In other words, the ARMA version of the RNN model in \eqref{Equ:DRNN:Model_hiddenEqu_contextwindow} can be implemented in an equivalent manner by having a context window that slides through several input samples and deliver them as an augmented input sample. From now on, we will proceed with the simple AR model \eqref{Equ:DRNN:Model_hiddenEqu}--\eqref{Equ:DRNN:Model_outputEqu} since the general ARMA version can be reduced to the AR version by the above conversion and hence the learning method will stay unchanged from the AR version to the ARMA version. Note that another approach to letting the model look into future is the bi-directional recurrent neural network\citep{graves2013speech}, which uses information from future by letting its hidden layers depend on the hidden states at future. On the other hand, the ARMA is made to depend on future by letting its input include future. So these two methods uses information from future in two different ways and our ARMA method is relatively easier to implement and also effectively capture the future information. Furthermore, by incorporating past and future information into the inputs, the ARMA-RNN also provides a relative easy way of enhancing the amount of memory in the RNN.

\subsection{Learning the RNN and its difficulties}
\label{Sec:DRNN:Difficulties}
The standard BPTT method for learning RNNs ``unfolds'' the network in time and propagates
error signals backwards through time.
Let $J(\Theta)$ denote the cost function that measures how good the RNN predicts
the target, where $\Theta \defeq \{ \W, \W_I, \U, \b\}$ is the set of all the parameters
to be trained. In general, the cost can be written as an average of costs over different
time instants:
	\begin{align}
		J(\Theta)	=	\frac{1}{T}\sum_{t=1}^T J_t(\y_t, \d_t)
		\label{Equ:DRNN:J_def}
	\end{align}
where $\d_t$ denotes the desired signal (also called target) at time $t$, and $J_t(\y_t,\d_t)$ characterizes the cost at time $t$. The cost $J_t(\y_t,\d_t)$ depends on the model parameters $\Theta$ through $\y_t$, which, as shown in \eqref{Equ:DRNN:Model_hiddenEqu}--\eqref{Equ:DRNN:Model_outputEqu}, further depends on $\{\W,\W_I,\U, \b\}$. Typical choices of $J_t(\y_t, \d_t)$ includes squared-error and cross entropy, etc. The gradient formula for $J(\Theta)$ with respect to $\Theta \defeq \{ \W, \W_I, \U, \b\}$ can be computed by back propagation through time (BPTT), which are given in sections \ref{Appendix:FormulaBPTT}--\ref{Appendix:Proof_BPTT} of the supplementary material.

It is well known that training RNNs is difficult
because of the exploding and vanishing gradient problems  \citep{pascanu2012difficulty}. Let 
	\begin{align}
		\gamma	\defeq		\| \D_f \|
	\end{align}	 
where $\D_f$ is a diagonal matrix constructed from the element-wise derivative of $\f(\x)$:
	\begin{align}
		\D_f	\defeq	\diag\{ \f'(\p_t) \}
	\end{align}
and $\|\X\|$ denotes the $2-$norm (the largest singular value) of the matrix $X$. For sigmoid function $\f(\x)$, $\gamma = 1/4$, and for $\tanh$, $\gamma = 1$. The argument in \citep{pascanu2012difficulty} pointed out that a sufficient condition for vanishing gradient problem to occur is
	\begin{align}
		\| \W \| = \|\W^T\|	< 	\frac{1}{\gamma}
		\label{Equ:DRNN:VanishingGradient}
	\end{align}
and a necessary condition for exploding gradient to occur is that
	\begin{align}
		\| \W \| = \|\W^T\|	>	\frac{1}{\gamma}
		\label{Equ:DRNN:ExplodingGradient}
	\end{align}
Therefore, the properties of the recurrent matrix $\W$ is essential for the performance of the RNN. In previous work \citep{pascanu2012difficulty,mikolov2011strategies}, the way to solve the exploding gradient problem is to empirically clip the gradient if the norm of the gradient exceeds a certain threshold, and the way to avoid the vanishing gradient is also empirical --- either to add a regularization term to push up the gradient or exploit the information about the curvature of the objective function \citep{maas2012recurrent}. Below, we describe an alternative primal-dual approach for training RNN, which is based on the echo-state property, a sufficient condition for avoiding the exploding gradient problem.

\section{A New Algorithm for Learning the RNN}
\label{Sec:NewLearning}

In this section, we first discuss an essential property for the model of the kind in \eqref{Equ:DRNN:Model_hiddenEqu} --- the echo-state property, and provide a sufficient condition for it. Then, we formulate the problem of training the RNN that preserves the echo-state property as a constrained optimization problem. Finally, we develop a primal-dual method to solve the problem.

\subsection{The echo-state property}
\label{Sec:DRNN:ESN_Property}
We now show that conditions of the form \eqref{Equ:DRNN:VanishingGradient}--\eqref{Equ:DRNN:ExplodingGradient} are closely related to whether the model \eqref{Equ:DRNN:Model_hiddenEqu}--\eqref{Equ:DRNN:Model_outputEqu} satisfies the echo-state property \citep{jaeger2002tutorial}, which states that ``if the network has been run for a very long time [from minus infinity time in the definition], the current network state is uniquely determined by the history of the input and the (teacher-forced) output''. It is also shown in \citep{jaeger2001short} that this echo-state property is  equivalent to the ``state contracting'' property. Since we do not consider the case with feedback from output in the model \eqref{Equ:DRNN:Model_hiddenEqu}, we here describe the ``state contracting'' property slightly different from \citep{jaeger2001short}:
	\begin{definition}[State contracting]
		\label{Def:StateContracting}
		The network is state contracting if for all right-infinite input sequences
		$\{\v_t\}$, where $t = 0,1,2,\ldots$, there exists a null sequence $(\epsilon_t)_{t \ge 0}$
		such that for all starting states $\h_0$ and $\h_0'$ and for all $t > 0$ it holds that
		$\|\h_t - \h_t'\| < \epsilon_t$, where $\h_t$ [resp. $\h_t'$] is the hidden state vector 
		at time $t$ obtained when the network is driven by $\v_t$ up to time $t$ after having been 
		stated in $\v_0$, [resp. in $\h_0'$].
	\end{definition}

It is shown in \citep{jaeger2001echostate} that a sufficient condition for the \emph{non-existence}, i.e., a necessary condition for the \emph{existence}, of the echo-state property is that the spectral radius of the recurrent matrix $\W$ is greater than one (resp. four) when $\tanh$ (resp. sigmoid) units are used. And it is further conjectured that for a weight matrix $\W$ that is randomly generated by sampling the weights over uniform distribution over $[-1,1]$ and normalized such that $\rho(\W) = 1-\delta$ would satisfy the echo-state property with high probability, where $\delta$ is a small positive number. 

In echo-state networks, the reservoir or the recurrent weights  are randomly generated and normalized according to the rule above and will be fixed over time in the training. The  input weights are fixed as well. Instead, in this paper, \emph{we learn the recurrent weights together with the input and output weights subject to the constraint that the network satisfies the echo-state property.} To this end, we propose the following sufficient condition for the echo-state property, which will be shown to be easily handled in the training procedure. The proof of the following proposition can be found in Appendix \ref{Appendix:Proof_SufficientESP}.
	\begin{proposition}[Echo-state: a sufficient condition]
		\label{Prop:ESN_sufficient}
		Let $\gamma \defeq \max_{x}  |f'(x)|$.
		The network model \eqref{Equ:DRNN:Model_hiddenEqu} satisfies the echo-state property if 
			\begin{align}
				\|\W\|_{\infty}		<	\frac{1}{\gamma}
				\label{Equ:Prop:ESN_sufficient}
			\end{align}
		where $\| \W \|_{\infty}$ denote the $\infty$-norm of the matrix $\W$ 
		(maximum absolute row sum), and $\gamma=1$ for $\tanh$ units, and
		$\gamma=1/4$
		for sigmoid units.
	\end{proposition}
	
The echo state property is typically assumed to capture a short-term memory in the data. However, it is reasonable for many applications, such as speech recognition, using RNNs since each phone typically lasts for fewer than 60 frames, requiring relatively short memory. Suppose we have two RNNs that run on different input sequences up to some time t so that they have different hidden states at time $t$. Afterwards, if the input acoustic sequences to the two RNNs become identical, the echo state property requires that the hidden states of these two RNNs be close to each other in order to produce the same predicted labels. Another important consequence of condition \eqref{Equ:Prop:ESN_sufficient} is that it also provides a sufficient condition to avoid the exploding gradient problem in a more principled manner. This can be shown by following the same argument as \citep{pascanu2012difficulty} except that the $2$-norm is replaced by $\infty$-norm. Thus, if the condition \eqref{Equ:Prop:ESN_sufficient} can be enforced in the training process, there is no need to clip the gradient. We will show in next section that the $\infty$-norm constraint on $\W$ is convenient to handle in optimization. Therefore, our proposed method below provides a relatively easy way to train RNN in a principled manner, especially on tasks that requires short memory. Learning RNN under refined characterization of its dynamics, such as \citep{manjunath2013echo}, is left as future work.

\subsection{Formal formulation of the learning problem}
\label{Sec:NewLearning:Form}
Based on the previous discussion, we can formulate the problem of training the RNN that preserves the echo-state property as the following constrained optimization problem:
	\begin{align}
		\min_{\Theta}	&	\quad J(\Theta) = J(\W, \W_I, \U, \b)			\\
		\mathrm{s.t.}	&	\quad \|\W\|_{\infty} \le \frac{1}{\gamma}
	\end{align}
In other words, we need to find the set of model parameters that best predict the target while preserving the echo-state property. Recall that $\|\W\|_{\infty}$ is defined as the maximum absolute row sum. Therefore, the above optimization problem is equivalent to the following constrained optimization problem (since $\max\{x_1,\ldots, x_N\} \le 1/\gamma$ is equivalent to $x_i \le 1/\gamma, \; i=1,\ldots,N$):
	\begin{align}
		\min_{\Theta}	&	\quad J(\Theta) = J(\W, \W_I, \U, \b)	
		\label{Equ:NewLearn:Optimization_Objective}		
							\\
		\mathrm{s.t.}	&	\quad \sum_{j=1}^N |W_{ij}|	\le 	\frac{1}{\gamma}, \quad i = 1,\ldots,N
		\label{Equ:NewLearn:Optimization_Constraint}
	\end{align}
where $W_{ij}$ denotes the $(i,j)$-th entry of the matrix $\W$. Next, we will proceed to derive the learning algorithm that can achieve this objective.

\subsection{Primal-dual method for optimizing RNN parameters}
\label{Sec:NewLearning:PD}
We solve the constrained optimization problem \eqref{Equ:NewLearn:Optimization_Objective}--\eqref{Equ:NewLearn:Optimization_Constraint} by primal-dual method. The Lagrangian of the problem can be written as
	\begin{align}
		\mc{L}(\Theta, \bm{\lambda})
			=	J(\W, \W_I, \U, \b) 
				\!+\! 
				\sum_{i=1}^N \lambda_i \left(\sum_{j=1}^N |W_{ij}| \!-\! \frac{1}{\gamma}\right)
		\label{Equ:NewLearn:Lagrangian}
	\end{align}
where $\lambda_i$ denotes the $i$th entry of the Lagrange vector $\bm{\lambda}$ (or dual variable) and is required to be non-negative. Let the dual function $q(\bm{\lambda})$ be defined as the following \emph{unconstrained} optimization problem:
	\begin{align}
		q(\bm{\lambda})	=	\min_{\Theta} \mc{L}(\Theta,\bm{\lambda})
		\label{Equ:NewLearn:DualFunc_def}
	\end{align}
It is shown that the dual function $q(\bm{\lambda})$ obtained from the above unconstrained optimization problem is always concave, even when the original cost $J(\Theta)$ is not convex \citep{boyd2004convex}. And the dual function is always a lower bound of the original constrained optimization problem \eqref{Equ:NewLearn:Optimization_Objective}--\eqref{Equ:NewLearn:Optimization_Constraint}: $q(\bm{\lambda}) \le J(\Theta^{\star})$. Maximizing $q(\bm{\lambda})$ subject to the constraint $\lambda_i \ge 0$, $i=1,\ldots, N$ will be the best lower bound that can be obtained from the dual function \citep{boyd2004convex}. This new problem is called the dual problem of the original optimization problem \eqref{Equ:NewLearn:Optimization_Objective}--\eqref{Equ:NewLearn:Optimization_Constraint}:
	\begin{align}
		\max_{\bm{\lambda}}	&	\quad q(\bm{\lambda})
		\label{Equ:NewLearn:OptimizationDual_Objective}		
							\\
		\mathrm{s.t.}	&	\quad \lambda_i \ge 0, \quad i = 1,\ldots,N
		\label{Equ:NewLearn:OptimizationDual_Constraint}
	\end{align}
which is a convex optimization problem since we are maximizing a concave objective with linear inequality constraints. After solving $\bm{\lambda}^{\star}$ from \eqref{Equ:NewLearn:OptimizationDual_Objective}--\eqref{Equ:NewLearn:OptimizationDual_Constraint}, we can substitute the corresponding $\bm{\lambda}^{\star}$ into the Lagrangian \eqref{Equ:NewLearn:Lagrangian} and then solve the correponding set of parameters $\Theta^o=\{\W^o, \W_I^o, \U^o, \b^o\}$ that minimizes $\mc{L}(\Theta,\bm{\lambda})$ for this given $\bm{\lambda}^{\star}$:
	\begin{align}
		\Theta^o		=	\arg\min_{\Theta} \mc{L}(\Theta, \bm{\lambda}^{\star})
	\end{align}
Then, the obtained $\Theta^o=\{\W^o, \W_I^o, \U^o, \b^o\}$ will be an approximation to the optimal solutions. In convex optimization problems, this approximate solution will be the same global optimal solution under some mild conditions \citep{boyd2004convex}. This property is called strong duality. However, in general non-convex problems, it will not be the exact solution. But since finding the globally optimal solution to the original problem \eqref{Equ:NewLearn:Optimization_Objective}--\eqref{Equ:NewLearn:Optimization_Constraint} is not realistic, it would be satisfactory if it can provide a good approximation. 

Back to the problem \eqref{Equ:NewLearn:OptimizationDual_Objective}--\eqref{Equ:NewLearn:OptimizationDual_Constraint}, we are indeed solving the following problem
	\begin{align}
		\max_{\bm{\lambda} \succeq \bm{0}} \min_{\Theta} \mc{L}(\Theta, \bm{\lambda})
	\end{align}
where the notation $\bm{\lambda} \succeq \bm{0}$ means that each entry of the vector $\bm{\lambda}$ is greater than or equal to zero. \emph{Now, to solve the problem, what we need to do is to minimize the Lagrangian $\mc{L}(\Theta, \bm{\lambda})$ with respect to $\Theta$, and in the mean time, maximize the dual variable $\bm{\lambda}$ subjected to the constraint that $\bm{\lambda} \succeq \bm{0}$}. Therefore, as we will see soon, updating the RNN parameters consists of two steps: \emph{primal update} (minimization of $\mc{L}$ with respect to $\Theta$) and \emph{dual update} (maximization of $\mc{L}$ with respect to $\bm{\lambda}$).

First, we provide the primal update rule.
To minimize the Lagrangian $\mc{L}(\Theta,\bm{\lambda})$ with respect to $\Theta$, we may apply gradient descent to $\mc{L}(\Theta,\bm{\lambda})$ with respect to $\Theta$. However, note that $\mc{L}(\Theta,\bm{\lambda})$ consists of two parts: $J(\Theta)$ that measures the prediction quality, and the part that penalizes the violation of the constraint \eqref{Equ:NewLearn:Optimization_Constraint}. Indeed, the second part is a sum of many $\ell_1$ regularization terms on the rows of the matrix $\W$:
	\begin{align}
		\sum_{j=1}^N |W_{ij}|	=	\| \w_i \|_{1}
	\end{align}
where $\w_i$ denotes the $i$th row vector of the matrix $\W$. With such observations, the Lagrangian in \eqref{Equ:NewLearn:Lagrangian} can be written in the following equivalent form:
	\begin{align}
		\mc{L}(\Theta, \bm{\lambda})
			=	J(\W, \W_I, \U, \b) + \sum_{i=1}^N \lambda_i \left(\|\w_i\|_1 - \frac{1}{\gamma}\right)
	\end{align}
To Minimize $\mc{L}(\Theta,\bm{\lambda})$ of the above structure with respect to $\Theta = \{\W,\W_I,\U, \b\}$, we can apply an argument similar to the one made in \citep{beck2009fast} to derive the following iterative soft-thresholding algorithm for the primal update of $\W$:
	\begin{align}
		\W_k		&=	\mc{T}_{\bm{\lambda} \mu_k}\!
					\left\{
						\W_{k-1} \!-\! \mu_k \frac{\partial J(\W_{k\!-\!1},\W_{I,{k\!-\!1}},\U_{k\!-\!1}, \b_{k\!-\!1})}{\partial \W}
					\right\}
		\label{Equ:NewLearn:primalUpdate_W}
	\end{align}
where $\mc{T}_{\bm{\lambda} \mu_k}(\X)$ denote a component-wise shrinkage (soft-thresholding) operator on a matrix $\X$, defined as
	\begin{align}
		[\mc{T}_{\bm{\lambda} \mu_k}(\X)]_{ij}	
				=	\begin{cases}
						X_{ij} - \lambda_i \mu_k	&	X_{ij} \ge \lambda_i \mu_k			\\
						X_{ij} + \lambda_i \mu_k	&	X_{ij} \le -\lambda_i \mu_k			\\
						0						& 	\mathrm{otherwise}
					\end{cases}
	\end{align}
Note that the primal update for $\W$ is implemented by a standard (stochastic) gradient descent followed by a shrinkage operator.
On the other hand, the primal updates for $\W_I$, $\U$ and $\b$ follow the standard (stochastic) gradient descent rule:
	\begin{align}
		\W_{I,k}	&=	\W_{I,k-1} - \mu_k \frac{\partial J(\W_{k-1},\W_{I,k-1},\U_{k-1}, \b_{k-1})}{\partial \W_I}
		\label{Equ:NewLearn:primalUpdate_WI}
					\\
		\U_{k}		&=	\U_{k-1} - \mu_k \frac{\partial J(\W_{k-1},\W_{I,k-1},\U_{k-1}, \b_{k-1})}{\partial \U}
		\label{Equ:NewLearn:primalUpdate_U}
					\\
		\b_k			&=	\b_{k-1} - \mu_k \frac{\partial J(\W_{k-1},\W_{I,k-1},\U_{k-1}, \b_{k-1})}{\partial \b}
		\label{Equ:NewLearn:primalUpdate_b}
	\end{align}
In order to accelerate the convergence of the algorithm, we can, for example, add momentum or use Nesterov method \citep{sutskever2013importance} to replace the gradient descent steps in \eqref{Equ:NewLearn:primalUpdate_W}--\eqref{Equ:NewLearn:primalUpdate_U}. In our experiments reported in Section 5, we adopted Nesterov method to accelerate the training process.

Next, we describe the dual update, which aims to maximize $\mc{L}$ with respect to $\bm{\lambda}$ subject to the constraint that $\bm{\lambda}\succeq \bm{0}$. To this end, we use the following rule of gradient ascent with projection, which increases the function value of $\mc{L}$ while enforcing the constraint:
	\begin{align}
		\lambda_{i,k}	=	\left[
								\lambda_{i,k-1} 
								+ 
								\mu_k 
								\left(
									\|\w_{i,k-1}\|_1 - \frac{1}{\gamma}
								\right)
							\right]_{+}
		\label{Equ:NewLearn:DualUpdate}
	\end{align}
where $[x]_{+} \defeq \max\{0,x\}$. Note that $\lambda_i$ is indeed a regularization factor in $\mc{L}(\Theta,\bm{\lambda})$ that penalizes the violation of constraint \eqref{Equ:NewLearn:Optimization_Constraint} for the $i$th row of $\W$. The dual update can be interpreted as a rule to adjust the regularization factor in an adaptive manner. When the sum of the absolute values of the $i$th row of $\W$ exceeds $1/\gamma$, i.e., violating the constraint, the recursion \eqref{Equ:NewLearn:DualUpdate} will increase the regularization factor $\lambda_{i,k}$ on the $i$th row in \eqref{Equ:NewLearn:Lagrangian}. On the other hand, if the constraint \eqref{Equ:NewLearn:Optimization_Constraint} for a certain $i$ is not violated, i.e., $\|\w_i\|_1 < 1/\gamma$, then the dual update \eqref{Equ:NewLearn:DualUpdate} will decrease the value of the corresponding $\lambda_i$ so that $\|\w_i\|_1 - 1/\gamma$ is less penalized in \eqref{Equ:NewLearn:Lagrangian}. This process will repeat itself until the constraints \eqref{Equ:NewLearn:Optimization_Constraint} are satisfied. The projection operator $[x]_{+}$ makes sure that once the regularization factor $\lambda_i$ is decreased below zero, it will be set to zero and the constraint for the $i$th row in \eqref{Equ:NewLearn:OptimizationDual_Constraint} will not be penalized in \eqref{Equ:NewLearn:Lagrangian}. An alternative choice to enhance the constraint is to apply projection operator to $\mathbf{W}$ after each stochastic gradient descent update \eqref{Equ:NewLearn:primalUpdate_WI}--\eqref{Equ:NewLearn:primalUpdate_b}.


\section{Experiments and Results }
\label{Sec:ExperimentSetup}


\begin{table*}
   \centering
    \caption{Phone recognition error (percent) on the TIMIT core test set using DNN-top feature sequences as the input to the RNNs. The results are shown as a function
of two hyperparameters: the size of the RNN's hidden layer and the moving-average order of the RNN. } 
\begin{tabular}{|c|c|c|c|c|c|c|c|c| }
            \hline
           HiddenSz& Order=0 &Order=2   & Order =4  & Order=6  &Order=8   & Order=10  & Order=12 \ \\
            \hline \hline
            100       &  19.85     & 19.83      &  19.80      &    19.65   &  19.50      &  19.44       & 19.42        \  \\
            200       &  19.86     & 19.72      &  19.65      &    19.60   &  19.45      &  19.35       & 19.31        \  \\
            300       &  20.02     & 19.72      &  19.60      &    19.56   &  19.40      &  19.23       & 19.16        \  \\
            500       &  20.00     & 19.73      &  19.56      &    19.44   &  19.34      &  19.06       & 18.91        \  \\
          1000       &  20.44     & 19.83      &  19.60      &    19.49   &  19.24      &  19.10       & 18.98        \  \\
          2000       &  20.70     & 20.34      &  20.10      &    19.65   &  19.45      &  19.30       & 19.12        \  \\
            \hline
        \end{tabular}
    \label{tab:Accuracy}
\end{table*}

\begin{table*}
\renewcommand{\arraystretch}{1.3}
	\centering
	\caption{Phone recognition error as a function of the clipping threshold (\citep{mikolov2011strategies}; \citep{pascanu2012difficulty}). The lowest error 19.05\% around threshold 1.0 is nevertheless higher than 18.91\% obtained by the new primal-dual method without tuning parameters.}
	\label{Tab:BPTT_gradclip}

	\begin{tabular}{|c|c|c|c|c|c|c|c|c|c|}
		\hline
		Threshold & 1 & 2 & 0.5 & 9 & 1.0 & 1.1 & 1.5 & 2 & 10 \\
		\hline
		Phone error (\%) & 19.65 & 19.50 & 19.25 & 19.10 & 19.05 & 19.08 & 19.15 & 19.54 & 20.5 \\
		\hline
	\end{tabular}
	
\end{table*}

We use the TIMIT phone recognition task to evaluate the deep-RNN architecture and the primal-dual optimization method for training the 
RNN part of the network.  The standard 462-speaker training set is used and all SA sentences are removed conforming to the standard protocol \citep{lee1989speaker,hinton2012deep,deng2013deep}. A separate development set of 50 speakers is used
for tuning all hyper parameters. Results are reported using the 24-speaker core
test set, which has no overlap with the development set. 


In our experiments, standard signal processing techniques are used for the raw speech waveforms, and 183 target class labels are used with three states for each of 61 phones.  After decoding, the original 61 phone classes are mapped to a set of 39 classes for final scoring according to the standard evaluation protocol.  In our experiments, a bi-gram language model over
phones, estimated from the training set, is used in decoding. To prepare the DNN and RNN targets, a high-quality tri-phone HMM model is trained on the training data set, which is then used to generate state-level labels based on HMM forced alignment. 


The DNN as part of the baseline described in \citep{deng2013deep} is used in this work to extract the high-level features from raw speech features. That is, the DNN features are discriminatively learned.
In our experiments, we use a DNN with three hidden layers, each having 2000 units. Each hidden layer's output vector, with a dimensionality of 2000, can be utilized as the DNN features. Thus, we have three sets of high-level features: DNN-top, DNN-middle, and DNN-bottom, indicated in Figure 1 with three separate colors of arrows pointing from DNN to RNN. 



We first report the phone recognition error performance of the deep-RNN using the DNN-top features. In Table \ref{tab:Accuracy}, 
the percent error results are shown as a function of the RNN hidden layer's size, varying from 100 to 2000, and also as a function of the moving average (MA) order in the RNN.   
Note when MA order is zero, the ARMA version of the RNN reverts to the traditional AR version. 

The above results are obtained using the fixed insertion penalty of zero and the fixed bi-phone ``language model'' weight of one.
When the language model's weight is tuned slightly over the development set, the core test set error further drops to 
18.86\%,
which is close to the best numbers reported recently
on this benchmark task in \citep{deng2013deep} using deep  convolutional neural networks with special design of the pooling strategy (18.70\%) and in \citep{graves2013speech} using a bi-directional RNN with special design of the memory structure (17.7\%). No bi-directionality and no special design on any structure are used in the RNN reported in this paper. 
The confusion matrix of this best result is shown in Section \ref{Appendix:ConfusionMatrix} of the supplementary document. 

We next compare the new primal-dual training method with the classical BPTT using gradient clipping as described in \citep{pascanu2012difficulty}. Table \ref{Tab:BPTT_gradclip} shows the the phone recognition error of the classical BPTT with gradient clipping on the TIMIT benchmark. We found that the error rate is sensitive to the threshold value. The best phone error rate on the test set is found to be between 19.05\%-20.5\% over a wide range of the threshold values where the best tuned clipping threshold is around 1.0 which corresponds to the error rate of 19.05\%. This is higher than the 18.91\% from our primal-dual method (without tuning the language model weights). Thus, using the new method presented in the paper, we do not need to tune the hyper-parameter of clipping threshold while obtaining lower errors.

We finally show the phone recognition error (percent) results for the features of DNN-middle, DNN-bottom, and of raw filter-bank data.
The size of the RNN's hidden layer is fixed at 500. And the results for four different MA orders are shown in Table \ref{tab:methods}.


\begin{table}
  \centering
    \caption{Phone recognition error (percent) on the TIMIT core test set using DNN-middle and DNN-bottom features, as well as the raw filter-bank features. } 
\begin{tabular}{|c|c|c|c|c| }
            \hline
         Features              & Order0 &Order4   & Order8  & Order12   \ \\
            \hline \hline
            DNN-Middle       &  20.70     & 20.30      &  19.96      &    19.65          \  \\
            DNN-Bottom      &  23.10     & 22.65      &  22.00      &    21.50          \  \\
            Filter-Banks       &  30.50     & 30.15      &  29.40      &    28.15          \  \\
            \hline
        \end{tabular}
    \label{tab:methods}
\end{table}

Comparisons among the results in Tables \ref{tab:Accuracy} and \ref{tab:methods} on phone recognition provide strong evidence that the high-level features extracted by DNNs are extremely useful for lowering recognition errors by the RNN. Further, the higher hidden layers in the DNN are more useful than the lower ones, given the same dimensionality of the features (fixed at 2000 as reported in this paper but we have the same conclusion for all other dimensions we have experimented). In addition, the introduction of MA components in the traditional AR version of the RNN also contributes significantly to reducing recognition errors.

\section{Discussion and Conclusion}
\label{Sec:Discussion}

The main contribution of the work described in this paper is a formulation of the RNN that lends itself to effective learning using a formal optimization
framework with a natural inequality constraint that provides a sufficient condition to guarantee the stability of the RNN dynamics during learning. This new learning method overcomes the
challenge of the earlier echo-state networks that typically fixed the recurrent weights due to the well-known difficulty in learning them. During the development of our new method, we propose a sufficient condition for the echo-state property, which is shown to be easily incorporated in the training procedure. We also make contributions to a new deep-RNN architecture, where the MA part is added to the original AR-only version of the RNN. The learning of both AR and ARMA versions of the RNN is unified after re-formulation of the model, and hence the same learning method developed can be applied to both versions.

The experimental contributions of this work are of four folds. First, we successfully apply the new learning method for the RNN to achieve close-to-record low error rates in phone recognition in the TIMIT benchmark. Second, we demonstrate the effectiveness of using DNNs to extract high-level features from the speech signal for providing the inputs to the RNN. With a combination of the DNN and RNN, we form a novel architecture of the deep-RNN, which, when trained separately, mitigates the overfitting problem. Third, on the same TIMIT benchmark task, we demonstrate clear superiority of the ARMA version of the RNN over the traditional AR version. The same efficient learning procedure is applied to both versions, with the only difference in the additional need to window the DNN outputs in the ARMA version of the RNN. Fourth, we show experimentally that the new training method motivated by optimization methodology achieves satisfactory performance as a sequence classifier on TIMIT benchmark task. Compared to the previous methods \citep{mikolov2011strategies,pascanu2012difficulty} of learning RNNs using heuristic rules of truncating gradients during the BPTT procedure,  our new method reports slightly lower phone recognition errors on the TIMIT benchmark and no longer needs to tune the threshold parameter as in the previous methods.

%
%


\bibliography{DeepRNN}

\begin{thebibliography}{36}
\providecommand{\natexlab}[1]{#1}
\providecommand{\url}[1]{\texttt{#1}}
\expandafter\ifx\csname urlstyle\endcsname\relax
  \providecommand{\doi}[1]{doi: #1}\else
  \providecommand{\doi}{doi: \begingroup \urlstyle{rm}\Url}\fi

\bibitem[Beck \& Teboulle(2009)Beck and Teboulle]{beck2009fast}
Beck, A. and Teboulle, M.
\newblock A fast iterative shrinkage-thresholding algorithm for linear inverse
  problems.
\newblock \emph{SIAM Journal on Imaging Sciences}, 2\penalty0 (1):\penalty0
  183--202, 2009.

\bibitem[Bengio et~al.(2013)Bengio, Boulanger-Lewandowski, and
  Pascanu]{bengio2013advances}
Bengio, Y., Boulanger-Lewandowski, N., and Pascanu, R.
\newblock Advances in optimizing recurrent networks.
\newblock In \emph{Proc. ICASSP}, Vancouver, Canada, May 2013.

\bibitem[Boyd \& Vandenberghe(2004)Boyd and Vandenberghe]{boyd2004convex}
Boyd, S.~P. and Vandenberghe, L.
\newblock \emph{{Convex Optimization}}.
\newblock Cambridge university press, 2004.

\bibitem[Dahl et~al.(2012)Dahl, Yu, Deng, and Acero]{Dahl2012}
Dahl, G., Yu, D., Deng, L., and Acero, A.
\newblock Context-dependent pre-trained deep neural networks for
  large-vocabulary speech recognition.
\newblock \emph{IEEE Trans. on Audio, Speech and Language Processing},
  20\penalty0 (1):\penalty0 30--42, jan 2012.

\bibitem[Dahl et~al.(2011)Dahl, Yu, Deng, and Acero]{dahl2011large}
Dahl, G.~E., Yu, D., Deng, L., and Acero, A.
\newblock {Large vocabulary continuous speech recognition with
  context-dependent DBN-HMMs}.
\newblock In \emph{Proc. IEEE ICASSP}, pp.\  4688--4691, Prague, Czech, May
  2011.

\bibitem[Dahl et~al.(2013)Dahl, Sainath, and Hinton]{dahl2013improving}
Dahl, G.~E., Sainath, T.~N., and Hinton, G.~E.
\newblock Improving deep neural networks for lvcsr using rectified linear units
  and dropout.
\newblock In \emph{Proc. ICASSP}, pp.\  8609--8613, Vancouver, Canada, May
  2013. IEEE.

\bibitem[Deng et~al.(1994)Deng, Hassanein, and Elmasry]{deng1994analysis}
Deng, L., Hassanein, K., and Elmasry, M.
\newblock Analysis of the correlation structure for a neural predictive model
  with application to speech recognition.
\newblock \emph{Neural Networks}, 7\penalty0 (2):\penalty0 331--339, 1994.

\bibitem[Deng et~al.(2013{\natexlab{a}})Deng, Abdel-Hamid, and
  Yu]{deng2013deep}
Deng, L., Abdel-Hamid, O., and Yu, D.
\newblock A deep convolutional neural network using heterogeneous pooling for
  trading acoustic invariance with phonetic confusion.
\newblock In \emph{Proc. IEEE ICASSP}, Vancouver, Canada, May
  2013{\natexlab{a}}.

\bibitem[Deng et~al.(2013{\natexlab{b}})Deng, Hinton, and
  Kingsbury]{deng2013new}
Deng, L., Hinton, G., and Kingsbury, B.
\newblock {New types of deep neural network learning for speech recognition and
  related applications: An overview}.
\newblock In \emph{Proc. IEEE ICASSP}, Vancouver, Canada, May
  2013{\natexlab{b}}.

\bibitem[Deng et~al.(2013{\natexlab{c}})Deng, Li, Huang, Yao, Yu, Seide,
  Seltzer, Zweig, He, Williams, Gong, and Acero]{DengMicrosoft}
Deng, L., Li, J., Huang, J.-T., Yao, K., Yu, D., Seide, F., Seltzer, M., Zweig,
  G., He, X., Williams, J., Gong, Y., and Acero, A.
\newblock Recent advances in deep learning for speech research at microsoft.
\newblock In \emph{Proc. ICASSP}, Vancouver, Canada, 2013{\natexlab{c}}.

\bibitem[Graves(2012)]{graves2012sequence}
Graves, A.
\newblock Sequence transduction with recurrent neural networks.
\newblock In \emph{Representation Learning Workshp, ICML}, 2012.

\bibitem[Graves et~al.(2006)Graves, Fern{\'a}ndez, Gomez, and
  Schmidhuber]{graves2006connectionist}
Graves, A., Fern{\'a}ndez, S., Gomez, F., and Schmidhuber, J.
\newblock Connectionist temporal classification: labelling unsegmented sequence
  data with recurrent neural networks.
\newblock In \emph{Proc. ICML}, pp.\  369--376, Pittsburgh, PA, June 2006. ACM.

\bibitem[Graves et~al.(2013)Graves, Mohamed, and Hinton]{graves2013speech}
Graves, A., Mohamed, A., and Hinton, G.
\newblock Speech recognition with deep recurrent neural networks.
\newblock In \emph{Proc. ICASSP}, Vancouver, Canada, May 2013.

\bibitem[Hinton et~al.(2012)Hinton, Deng, Yu, Dahl, Mohamed, Jaitly, Senior,
  Vanhoucke, Nguyen, Sainath, and Kingsbury]{hinton2012deep}
Hinton, G., Deng, L., Yu, D., Dahl, G.~E., Mohamed, A., Jaitly, N., Senior, A.,
  Vanhoucke, V., Nguyen, P., Sainath, T.~N., and Kingsbury, B.
\newblock {Deep neural networks for acoustic modeling in speech recognition:
  The shared views of four research groups}.
\newblock \emph{IEEE Signal Processing Magazine}, 29\penalty0 (6):\penalty0
  82--97, November 2012.

\bibitem[Jaeger(2001{\natexlab{a}})]{jaeger2001echostate}
Jaeger, H.
\newblock \emph{The ``echo state'' approach to analysing and training recurrent
  neural networks}.
\newblock GMD Report 148, GMD - German National Research Institute for Computer
  Science, 2001{\natexlab{a}}.

\bibitem[Jaeger(2001{\natexlab{b}})]{jaeger2001short}
Jaeger, H.
\newblock \emph{Short term memory in echo state networks}.
\newblock GMD Report 152, GMD - German National Research Institute for Computer
  Science, 2001{\natexlab{b}}.

\bibitem[Jaeger(2002)]{jaeger2002tutorial}
Jaeger, H.
\newblock \emph{Tutorial on training recurrent neural networks, covering BPPT,
  RTRL, EKF and the ``echo state network'' approach}.
\newblock GMD Report 159, GMD - German National Research Institute for Computer
  Science, 2002.

\bibitem[Kingsbury et~al.(2012)Kingsbury, Sainath, and
  Soltau]{kingsbury2012scalable}
Kingsbury, B., Sainath, T.~N., and Soltau, H.
\newblock Scalable minimum bayes risk training of deep neural network acoustic
  models using distributed hessian-free optimization.
\newblock In \emph{Proc. INTERSPEECH}, Portland, OR, September 2012.

\bibitem[LeCun(2012)]{lecun2012learning}
LeCun, Yann.
\newblock Learning invariant feature hierarchies.
\newblock In \emph{Proc. ECCV}, pp.\  496--505, Firenze, Italy, October 2012.
  Springer.

\bibitem[Lee \& Hon(1989)Lee and Hon]{lee1989speaker}
Lee, K.-F. and Hon, H.-W.
\newblock {Speaker-independent phone recognition using hidden Markov models}.
\newblock \emph{IEEE Transactions on Acoustics, Speech and Signal Processing,},
  37\penalty0 (11):\penalty0 1641--1648, November 1989.

\bibitem[Maas et~al.(2012)Maas, Le, O'Neil, Vinyals, Nguyen, and
  Ng]{maas2012recurrent}
Maas, A.~L., Le, Q., O'Neil, T.~M., Vinyals, O., Nguyen, P., and Ng, A.~Y.
\newblock {Recurrent Neural Networks for Noise Reduction in Robust ASR.}
\newblock In \emph{Proc. INTERSPEECH}, Portland, OR, September 2012.

\bibitem[Manjunath \& Jaeger(2013)Manjunath and Jaeger]{manjunath2013echo}
Manjunath, G. and Jaeger, H.
\newblock {Echo state property linked to an input: Exploring a fundamental
  characteristic of recurrent neural networks}.
\newblock \emph{Neural computation}, 25\penalty0 (3):\penalty0 671--696, 2013.

\bibitem[Martens \& Sutskever(2011)Martens and Sutskever]{martens2011learning}
Martens, J. and Sutskever, I.
\newblock Learning recurrent neural networks with hessian-free optimization.
\newblock In \emph{Proc. ICML}, pp.\  1033--1040, Bellevue, WA, June 2011.

\bibitem[Mikolov(2012)]{mikolov2012statistical}
Mikolov, T.
\newblock \emph{Statistical Language Models Based on Neural Networks}.
\newblock PhD thesis, Ph. D. thesis, Brno University of Technology, 2012.

\bibitem[Mikolov et~al.(2011)Mikolov, Deoras, Povey, Burget, and
  Cernocky]{mikolov2011strategies}
Mikolov, T., Deoras, A., Povey, D., Burget, L., and Cernocky, J.
\newblock Strategies for training large scale neural network language models.
\newblock In \emph{Proc. IEEE ASRU}, pp.\  196--201, Honolulu, HI, December
  2011. IEEE.

\bibitem[Pascanu et~al.(2013)Pascanu, Mikolov, and
  Bengio]{pascanu2012difficulty}
Pascanu, R., Mikolov, T., and Bengio, Y.
\newblock On the difficulty of training recurrent neural networks.
\newblock In \emph{Proc. ICML}, Atlanta, GA, June 2013.

\bibitem[Polyak(1987)]{poliak1987introduction}
Polyak, B.
\newblock \emph{{Introduction to Optimization}}.
\newblock Optimization Software, NY, 1987.

\bibitem[Robinson(1994)]{robinson1994application}
Robinson, A.~J.
\newblock An application of recurrent nets to phone probability estimation.
\newblock \emph{IEEE Transactions on Neural Networks}, 5\penalty0 (2):\penalty0
  298--305, August 1994.

\bibitem[Sainath et~al.(2013)Sainath, Kingsbury, Soltau, and
  Ramabhadran]{sainath2013optimization}
Sainath, T.N., Kingsbury, B., Soltau, H., and Ramabhadran, B.
\newblock Optimization techniques to improve training speed of deep neural
  networks for large speech tasks.
\newblock \emph{IEEE Transactions on Audio, Speech, and Language Processing},
  21\penalty0 (11):\penalty0 2267--2276, November 2013.

\bibitem[Seide et~al.(2011)Seide, Li, and Yu]{seide2011conversational}
Seide, F., Li, G., and Yu, D.
\newblock Conversational speech transcription using context-dependent deep
  neural networks.
\newblock In \emph{Proc. INTERSPEECH}, pp.\  437--440, Florence, Italy, August
  2011.

\bibitem[Sutskever(2013)]{sutskever2013training}
Sutskever, I.
\newblock \emph{Training Recurrent Neural Networks}.
\newblock PhD thesis, Ph. D. thesis, University of Toronto, 2013.

\bibitem[Sutskever et~al.(2013)Sutskever, Martens, Dahl, and
  Hinton]{sutskever2013importance}
Sutskever, I., Martens, J., Dahl, G., and Hinton, G.~E.
\newblock On the importance of initialization and momentum in deep learning.
\newblock In \emph{Proc. ICML}, Atlanta, GA, June 2013.

\bibitem[T{\"u}ske et~al.(2012)T{\"u}ske, Sundermeyer, Schl{\"u}ter, and
  Ney]{Tuske12}
T{\"u}ske, Z., Sundermeyer, M., Schl{\"u}ter, R., and Ney, H.
\newblock {Context-Dependent MLPs for LVCSR: TANDEM, Hybrid or Both?}
\newblock In \emph{Proc. Interspeech}, Portland, OR, September 2012.

\bibitem[Vinyals et~al.(2012)Vinyals, Ravuri, and Povey]{vinyals2012revisiting}
Vinyals, Oriol, Ravuri, Suman~V, and Povey, Daniel.
\newblock Revisiting recurrent neural networks for robust {ASR}.
\newblock In \emph{Proc. ICASSP}, pp.\  4085--4088, Kyoto, Japan, March 2012.
  IEEE.

\bibitem[Yu et~al.(2010)Yu, Deng, and Dahl]{DongDengDahl}
Yu, D., Deng, L., and Dahl, G.
\newblock Roles of pre-training and fine-tuning in context-dependent {DBN-HMM}s
  for real-world speech recognition.
\newblock In \emph{NIPS Workshop on Deep Learning and Unsupervised Feature
  Learning}, 2010.

\bibitem[Yu et~al.(2013)Yu, Deng, and Seide]{Yu2013}
Yu, D., Deng, L., and Seide, F.
\newblock The deep tensor neural network with applications to large vocabulary
  speech recognition.
\newblock \emph{IEEE Trans. on Audio, Speech and Language Processing},
  21\penalty0 (2):\penalty0 388 --396, 2013.

\end{thebibliography}
\bibliographystyle{icml2014}

%
%
%
%

\newpage

\appendix

\section*{\huge Supplementary Material}

\section{Proof of a sufficient condition for echo-state property}
\label{Appendix:Proof_SufficientESP}

First, we prove the following property regarding $\f(\x)$:
	\begin{align}
		\|\f(\x) - \f(\y)\|_{\infty}	\le 	\gamma \cdot \| \x - \y \|_{\infty}
	\end{align}
where $\gamma \defeq \max f'(x)$. Let $x_k$ and $y_k$ denote the $k$-th entries of the $N\times 1$ vectors $\x$ and $\y$, respectively. Then, 
	\begin{align}
		\|\f(\x) - \f(\y)\|_{\infty}	&=		\max_{1 \le k \le N} \left| f(x_k) - f(y_k) \right|
		\label{Equ:Appendix:fxfy_difference_infmax}
	\end{align}
By mean value theorem, we have
	\begin{align}
		f(x_k)	&=	f(y_k) + \int_{0}^1 f'\big(y_k + t(x_k-y_k)\big) dt (x_k-y_k)
	\end{align}
so that
	\begin{align}
		|&f(x_k) - f(y_k)|			\nonumber\\
							&\le 	\left|
										\int_{0}^1 f'\big(y_k + t(x_k-y_k)\big) dt (x_k-y_k)
									\right|
									\nonumber\\
							&\le  	\int_{0}^1 \left|f'\big(y_k + t(x_k-y_k)\big)\right| dt 
									\cdot
									|x_k-y_k|
									\nonumber\\
							&\le 	\int_0^1 \gamma \cdot dt \cdot |x_k - y_k|
									\nonumber\\
							&\le 	\gamma \cdot |x_k-y_k|
	\end{align}
Substituting the above result into \eqref{Equ:Appendix:fxfy_difference_infmax}, we get
	\begin{align}
		\|\f(\x) - \f(\y)\|_{\infty}	&\le 	\gamma \cdot \max_{1 \le k \le N} |x_k - y_k|
												\nonumber\\
										&= 		\gamma \cdot \| \x - \y \|_{\infty}
		\label{Equ:Appendix:fxfy_vector_diff_infmax}
	\end{align}
From the expressions of $\tanh$ and sigmoid functions, we can easily verify that $\gamma = 1$ for $\tanh$ function and $\gamma = 1/4$ for sigmoid function.

Let $\h_t$ and $\h_t'$ denote the hidden states of the recurrent neural network \eqref{Equ:DRNN:Model_hiddenEqu} that starts at initial conditions $\h_0$ and $\h_0'$, respectively:
	\begin{align}
		\h_t	&=	\f\left( \W \h_{t-1} + \W_I \v_t + \b \right)
		\label{Equ:Appendix:ht_stateEqu}
					\\
		\h_t'	&=	\f\left( \W \h_{t-1}' + \W_I \v_t + \b \right)
		\label{Equ:Appendix:ht_prime_stateEqu}
	\end{align}
Subtracting \eqref{Equ:Appendix:ht_prime_stateEqu} from \eqref{Equ:Appendix:ht_stateEqu}, we obtain
	\begin{align}
		\h_t - \h_t'	&=	\f\left( \W \h_{t-1} + \W_I \v_t + \b \right)
							\nonumber\\
							&\quad
							-
							\f\left( \W \h_{t-1}' + \W_I \v_t + \b \right)
	\end{align}
Taking $\infty$-norm of both sides of the above expression and using \eqref{Equ:Appendix:fxfy_vector_diff_infmax}, we get
	\begin{align}
		\| &\h_t - \h_t' \|_{\infty}		\nonumber\\
						&=	\big\|
								\f\left( \W \h_{t-1} + \W_I \v_t + \b \right)
								\nonumber\\
								&\quad
								-
								\f\left( \W \h_{t-1}' + \W_I \v_t + \b \right)
							\big\|_{\infty}
							\nonumber\\
						&\le 
							\gamma \cdot
							\left\|
								\W (\h_{t-1} - \h_{t-1}')
							\right\|_{\infty}
							\nonumber\\
						&\le 
							\gamma
							\cdot
							\|\W\|_{\infty}
							\cdot
							\| \h_{t-1} - \h_{t-1}' \|_{\infty}
							\nonumber\\
						&\le 
							\big(
								\gamma
								\cdot
								\|\W\|_{\infty}
							\big)^{t}
							\cdot
							\| \h_{0} - \h_{0}' \|_{\infty}
	\end{align}
Let $\epsilon_t \defeq (\gamma	\cdot\|\W\|_{\infty})^{t} \cdot \| \h_{0} - \h_{0}' \|_{\infty}$. Obviously, as long as $\gamma \cdot \|\W\|_{\infty} < 1$ or, equivalently, $\|\W\|_{\infty} < 1/\gamma$, we have $\epsilon_t \rightarrow 0$ and we can conclude the network \eqref{Equ:DRNN:Model_hiddenEqu} is state contracting and thus has echo-state property.

\section{Gradient formula for BPTT}
\label{Appendix:FormulaBPTT}

Specifically, let $\p_t$ be a $N\times 1$ vector that collects the input to the hidden units:
	\begin{align}
		\p_t	&\defeq		\W \h_t + \W_I \v_t + \b	
		\label{Equ:DRNN:pt_def}	
	\end{align}
Let $\bm{\delta}_t$ denote the error signal to be propagated:
	\begin{align}
		\bm{\delta}_t	\defeq		T \cdot \frac{\partial J }{\partial \p_t}
						=			\frac{\partial}{\partial \p_t}
									\sum_{t=1}^T 
									J_t(\y_t, \d_t)
		\label{Equ:DRNN:delta_def}
	\end{align}
Then, by chain rule (see Section \ref{Appendix:Proof_BPTT} of the supplementary document for details), $\bm{\delta}_t$ is propagated through time according to the following backward recursion:
	\begin{align}
		\bm{\delta}_t	
			&=	\D_f \W^T \cdot \bm{\delta}_{t+1}
				+
				\frac{\partial J_t}{\partial \p_t},
				\quad t=T, T-1, \ldots, 0
		\label{Equ:DRNN:BPTT_delta_recursion}
	\end{align}
with initialization $\bm{\delta}_{T+1} = \mathbf{0}$, where $\D_f$ is a diagonal matrix constructed from the element-wise derivative of $\f(\x)$:
	\begin{align}
		\D_f	\defeq	\diag\{ \f'(\p_t) \}
	\end{align}
The form of $\frac{\partial J_t}{\partial \p_t}$ depends on the choice of cost functions $J_t$ and the output units $\g(\x)$. For example, for linear output units and square-error cost, i.e.,
	\begin{align}
		\g(\x)			&=	\x
							\\
		J_t(\y_t,\d_t) 	&= 	\| \y_t - \d_t \|^2
	\end{align} 
the expression for $\frac{\partial J_t}{\partial \p_t}$ is given by
	\begin{align}
		\frac{\partial J_t}{\partial \p_t}
				&=		-2 \U^T \left( \d_t - \y_t \right)
	\end{align}
On the other hand, if soft-max output units and cross-entropy cost are used, i.e., 
	\begin{align}
		\g(\x)_n			&=	\frac{e^{x_n}}{\sum_{n=1}^{N_o} e^{x_n}}
							\\
		J_t(\y_t,\d_t)	&=	-\sum_{n=1}^{N_o} d_{n,t} \ln y_{n,t}
	\end{align}
where $d_{n,t}$ and $y_{n,t}$ denote the $n$th entry of the vectors $\d_{t}$ and $\y_{t}$, respectively,
the expression for $\frac{\partial J_t}{\partial \p_t}$ will be given by
	\begin{align}
		\frac{\partial J_t}{\partial \p_t}
				&=		-(\d_t - \y_t)
	\end{align}
And the gradients of $J(\theta)$ with respect to the parameters $\W$, $\W_I$, $\U$ and $\b$
are given by the following expressions:
	\begin{align}
		\frac{\partial J}{\partial \W}	
				&=	\frac{1}{T} 
					\sum_{t=1}^T 
					\bm{\delta}_t \h_{t-1}^T
		\label{Equ:DRNN:W_gradient}
					\\
		\frac{\partial J}{\partial \W_I}	
				&=	\frac{1}{T} 
					\sum_{t=1}^T 
					\bm{\delta}_t \v_t^T
		\label{Equ:DRNN:WI_gradient}
					\\
		\frac{\partial J}{\partial \U}	
				&=	\frac{1}{T} 
					\sum_{t=1}^T 
					\frac{\partial J_t}{\partial\U}
		\label{Equ:DRNN:U_gradient}
					\\
		\frac{\partial J}{\partial \b}
				&=	\frac{1}{T}
					\sum_{t=1}^T
					\bm{\delta}_t
	\end{align}
For soft-max output units with cross-entropy cost, which we use in this work, $\frac{\partial J_t}{\partial \U}$ is given by
	\begin{align}
		\frac{\partial J_t}{\partial \U}		=		-(\d_t - \y_t) \h_t^T
	\end{align}

\section{Derivation of the back propagation recursion}
\label{Appendix:Proof_BPTT}
From the definition of the error signal in \eqref{Equ:DRNN:delta_def}, we can write $\bm{\delta}_t$ as
	\begin{align}
		\bm{\delta}_t					&=	\frac{\partial}{\partial \p_t} 
											\sum_{u=1}^T
											J_u
											\nonumber\\
										&=	\sum_{u=1}^T
											\frac{\partial J_u}{\partial \p_t} 
	\end{align}
In RNN, only $J_u$ with $u \ge t$ depend on $\p_t$, so we have $\frac{\partial J_u}{\partial \p_t} = 0$ for $ u < t$ and hence
	\begin{align}
		\frac{\partial J}{\partial \p_t}	&=	\sum_{u=t}^T
											\frac{\partial J_u}{\partial \p_t} 
											\nonumber\\
										&=	\sum_{u=t+1}^T
											\frac{\partial J_u}{\partial \p_t} 
											+
											\frac{\partial J_t}{\partial \p_t}
											\nonumber\\
										&=	\sum_{u=t+1}^T
											\frac{\partial \p_t^T}{\partial \p_{t+1}}
											\cdot
											\frac{\partial J_u}{\partial \p_{t+1}} 
											+
											\frac{\partial J_t}{\partial \p_t}
											\nonumber\\
										&=	\frac{\partial \p_t^T}{\partial \p_{t+1}}
											\cdot
											\sum_{u=t+1}^T											
											\frac{\partial J_u}{\partial \p_{t+1}} 
											+
											\frac{\partial J_t}{\partial \p_t}
											\nonumber\\
										&=	\frac{\partial \p_t^T}{\partial \p_{t+1}}
											\cdot
											\sum_{u=1}^T											
											\frac{\partial J_u}{\partial \p_{t+1}} 
											+
											\frac{\partial J_t}{\partial \p_t}
											\nonumber\\
										&=	\frac{\partial \p_t^T}{\partial \p_{t+1}}
											\cdot
											\bm{\delta}_{t+1}
											+
											\frac{\partial J_t}{\partial \p_t}
		\label{Equ:Appendix:backward_recursion_interm1}
	\end{align}
Now we evaluate $\frac{\partial \p_t^T}{\partial \p_{t+1}}$, which by chain rule, can be written as
	\begin{align}
		\frac{\partial \p_t^T}{\partial \p_{t+1}}
					&=			\frac{\partial \h_t^T}{\partial \p_t}
								\cdot
								\frac{\partial \p_{t+1}^T}{\partial \h_t}
		\label{Equ:Appendix:partial_pt_partial_pt1_interm1}
	\end{align}
By the expression of $\p_t$ in \eqref{Equ:DRNN:pt_def}, we have
	\begin{align}
		\frac{\partial \p_{t+1}^T}{\partial \h_t}
					&=			\frac{\partial}{\partial \h_t}
								\left\{
									\W \h_t + \W_I \v_t + \b	
								\right\}^T
								\nonumber\\
					&=			\W^T
		\label{Equ:Appendix:partial_pt_partial_pt1}
	\end{align}
and by \eqref{Equ:DRNN:Model_hiddenEqu}, we have
	\begin{align}
		\frac{\partial \h_t^T}{\partial \p_t}
					&=			\frac{\partial}{\partial \p_t}
								\left\{
									\f(\W \h_{t-1} + \W_I \v_t + \b)
								\right\}^T
								\nonumber\\
					&=			\frac{\partial}{\partial \p_t}
								\left\{
									\f(\p_t)
								\right\}^T
								\nonumber\\
					&=			\f'(\p_t)
								\nonumber\\
					&=			\D_f
		\label{Equ:Appendix:partial_ht_partial_pt}
	\end{align}
Substituting \eqref{Equ:Appendix:partial_pt_partial_pt1}--\eqref{Equ:Appendix:partial_ht_partial_pt} into \eqref{Equ:Appendix:partial_pt_partial_pt1_interm1}, we get
	\begin{align}
		\frac{\partial \p_t^T}{\partial \p_{t+1}}
					&=			\D_f \W^T
		\label{Equ:Appendix:partial_pt_partial_pt1_final}
	\end{align}
Substituting \eqref{Equ:Appendix:partial_pt_partial_pt1_final} into \eqref{Equ:Appendix:backward_recursion_interm1}, we conclude our proof of \eqref{Equ:DRNN:BPTT_delta_recursion}. Next, we derive \eqref{Equ:DRNN:W_gradient}--\eqref{Equ:DRNN:U_gradient}. We will only show the proof of \eqref{Equ:DRNN:W_gradient}. By chain rule, we have
	\begin{align}
		\frac{\partial J}{\partial W_{ij}}
					&=			\frac{1}{T}
								\cdot
								\frac{\partial}{\partial W_{ij}}
								\left(
									\sum_{u=1}^T
									J_u
								\right)
								\nonumber\\
					&=			\frac{1}{T}
								\cdot
								\sum_{t=1}^T
								\frac{\partial \p_t^T}{\partial W_{ij}}
								\cdot
								\frac{\partial}{\partial \p_t}
								\left(
									\sum_{u=1}^T
									J_u
								\right)
								\nonumber\\
					&=			\frac{1}{T}
								\sum_{t=1}^T
								\frac{\partial \p_t^T}{\partial W_{ij}}
								\bm{\delta}_t
	\end{align}
By \eqref{Equ:DRNN:pt_def}, we have
	\begin{align}
		\left[\frac{\partial \p_t^T}{\partial W_{ij}}\right]_{n}
					=			\begin{cases}
									[\h_{t}]_j	&	n=i		\\
									0			&	n \neq i
								\end{cases}
	\end{align}
where the notatin $[\x]_i$ denotes the $i$th entry of the vector. Therefore, 
	\begin{align}
		\frac{\partial J}{\partial W_{ij}}
					&=			\frac{1}{T}
								\sum_{t=1}^T
								[\h_t]_j [\bm{\delta}_t]_i
	\end{align}
which implies that, in matrix form, 
	\begin{align}
		\frac{\partial J}{\partial \W}
					&=			\frac{1}{T}
								\sum_{t=1}^T
								\bm{\delta}_t \h_t^T
	\end{align}

\section{Confusion matrix of phone recognition}
\label{Appendix:ConfusionMatrix}

See the figure in next page.

\begin{landscape}
\begin{figure}
	\includegraphics[width=1.7\textwidth]{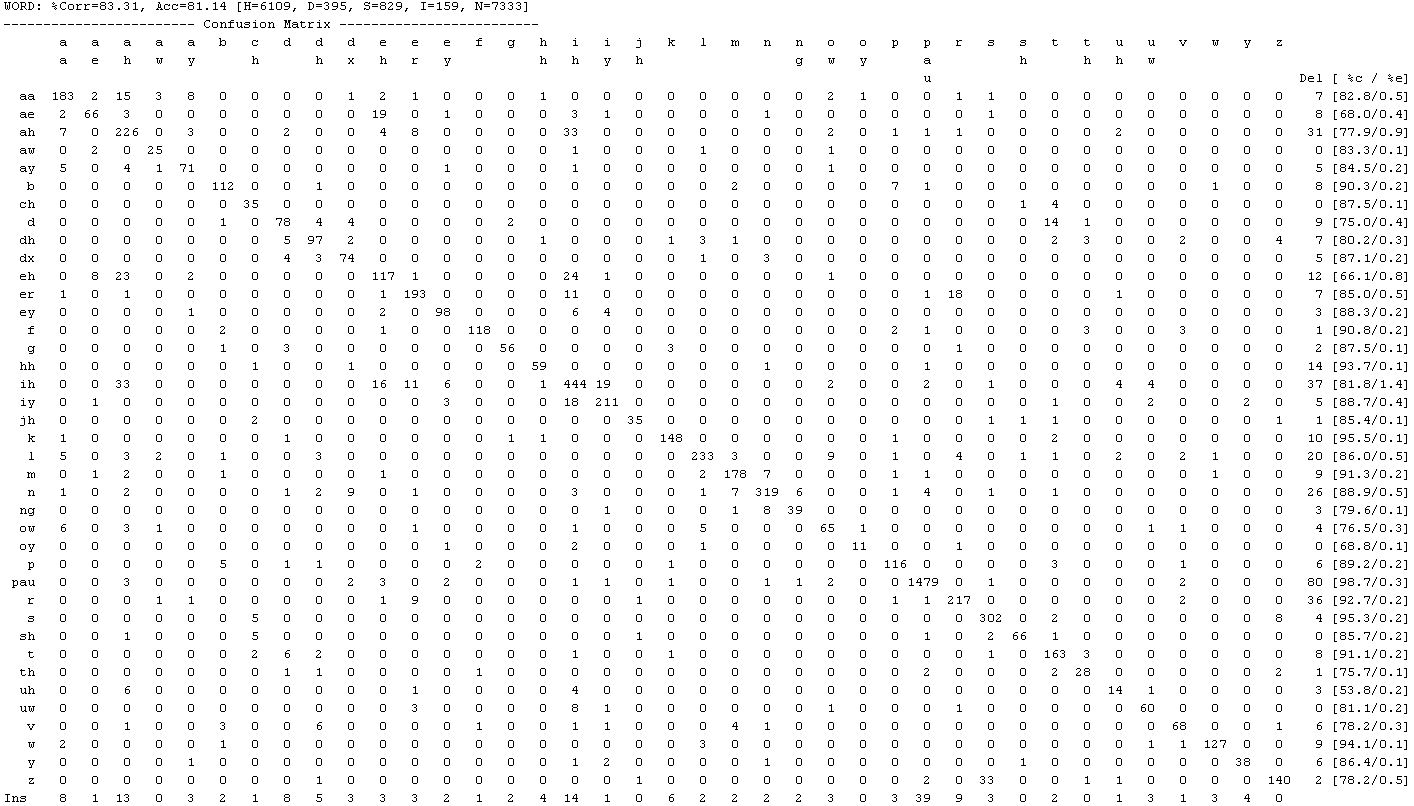}
\end{figure}

\end{landscape}

\end{document}